\documentclass[lettersize,journal]{IEEEtran}
\usepackage{amsmath,amsfonts}
\usepackage{algorithmic}
\usepackage{array}
\usepackage{textcomp}
\usepackage{stfloats}
\usepackage{url}
\usepackage{verbatim}
\usepackage{graphicx}
\usepackage{subcaption}

\usepackage{multirow}
\usepackage{booktabs}
\usepackage{tabularx}
\usepackage[table]{xcolor}
\usepackage{colortbl}
\usepackage{xfp}

\definecolor{tablegreen}{RGB}{0, 150, 0}
\definecolor{tableorange}{RGB}{255, 165, 0}
\definecolor{tablered}{RGB}{255, 0, 0}
\def\BibTeX{{\rm B\kern-.05em{\sc i\kern-.025em b}\kern-.08em
    T\kern-.1667em\lower.7ex\hbox{E}\kern-.125emX}}
\usepackage{balance}

\newcommand{\val}[1]{%
  \ifdim #1 pt > 95 pt
    \textcolor{green!60!black}{#1}%
  \else
    \ifdim #1 pt > 70 pt
      \textcolor{orange!85!black}{#1}%
    \else
      \textcolor{red!75!black}{#1}%
    \fi
  \fi
}

\begin{document}
\title{FEM-Validated Real-Time Modelling of Soft Tactile Deformation}

 \author{Silvia Terrile$^{1}$, Kaichen Shi$^{1}$, Taejun Kwon$^{2} $, Jiseok Lee$^{3}$ , Nathan Lepora$^{1}$, Yeejin Lee$^{3} $, Saekwang Nam$^{2} $ and Efi Psomopoulou$^{1}$
 \thanks{This work was supported by the Royal Society International Collaboration Award (South Korea) (No. ICA\textbackslash R1\textbackslash 241213) of  and a National Research Foundation of Korea (NRF) grant funded by the Korea government (MSIT) (No. RS-2024-00436182).}
 \thanks{$^{1}$School of Engineering Mathematics and Technology, University of Bristol, UK.
         {\tt\small [silvia.terrile, vr24150, n.lepora, efi.psomopoulou]@bristol.ac.uk}}%
 \thanks{$^{2}$Graduate School of Data Science, Kyungpook National University, Daegu, South Korea.
       {\tt\small [ktjmamamoo0629, s.nam]@knu.ac.kr}}
       
 \thanks{$^{3}$Department of Electrical and Information Engineering, Seoul National University of Science and Technology, Seoul, Republic of Korea. 
       {\tt\small [desiderio1971, yeejinlee]@seoultech.ac.kr}}%
        }
\markboth{Journal of \LaTeX\ Class Files,~Vol.~18, No.~9, September~2020}{Journal of \LaTeX\ Class Files,~Vol.~18, No.~9, September~2020}%

\maketitle

\begin{abstract}
Accurate modelling of soft tactile sensors remains difficult because finite element methods (FEM) are computationally expensive, while real-time simulators often lack fidelity for hyperelastic deformation. Large physical datasets are also impractical to collect because repeated testing can damage fragile sensors. We present a data-efficient framework that combines limited real-world validation, high-fidelity FEM, and graph neural networks (GNNs) for accurate real-time prediction of two-dimensional tactile deformation in a planar soft tactile structure. The key idea is to decouple validation from training: 82 camera-tracked contact configurations are used only to tune and validate an ANSYS FEM model, which then generates 768 synthetic contact scenarios for GNN training. We evaluate both forward prediction from marker deformation to contact localisation and inverse inference from contact configuration to marker deformation. The validated FEM achieves sub-0.5 mm error in more than 90\% of validation cases, while the learned GNN surrogate reaches a mean absolute error of 0.32 mm with millisecond-scale inference, compared with minute-scale FEM simulation. Models trained entirely on synthetic FEM data retain sub-millimetre accuracy on real measurements from the 2D tactile structure, demonstrating strong sim-to-real transfer within the tested planar setting. We further integrate the learned model into a real-time visualisation pipeline, enabling rapid deformation prediction for virtual prototyping and simulator-based design exploration without extensive physical testing.

\end{abstract}

\begin{figure}[!t]
\centering
\includegraphics[width=3.5in]{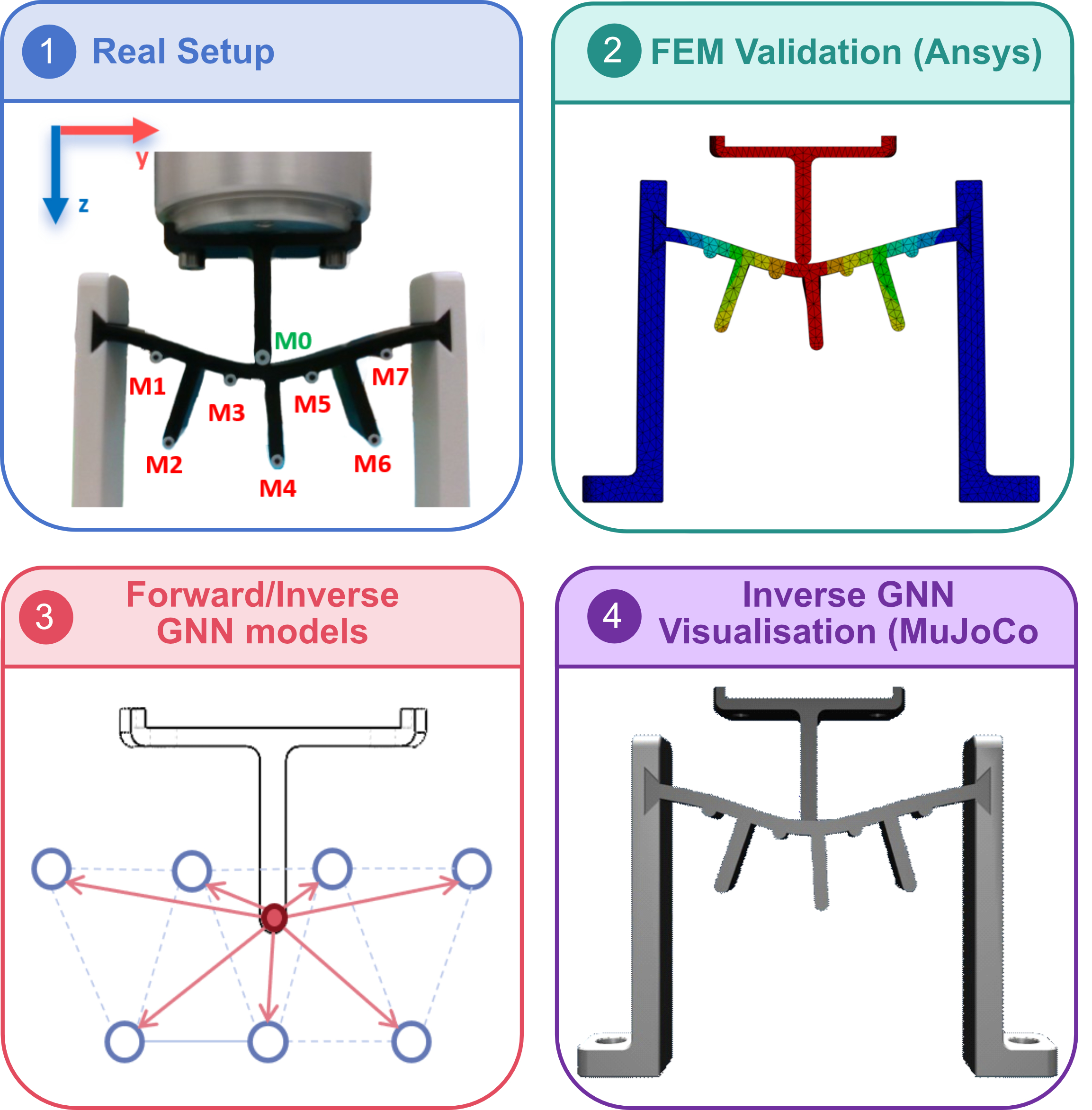}
\caption{(1) Experimental setup of the tactile sensing system
    with a compliant TPU pin structure constrained between
    rigid supports and loaded to an indentation depth of 14
    mm. Marker M0 denotes the indenter marker, while markers
    M1–M7 denote the pin markers, identifiable as white circular
    markers. (2) Corresponding ANSYS simulation showing the
    total deformation under identical loading conditions. (3) The
    trained Graph Neural Network facilitates real-time inference
    of structural deformation based on the spatial coordinates of
    the indenter. (4) Visualisation of the structural deformation in a real-time robotic simulator. }
\label{fig:stacked}
\end{figure}

\section{Introduction}
\IEEEPARstart{S}{oft} tactile sensors play a key role in robotic manipulation by providing the fine-grained contact feedback needed for dexterous tasks. Yet, modelling how these sensors deform under contact is both computationally demanding and constrained by the delicate nature of soft materials. Finite element methods (FEM) offer high accuracy but are too slow for large datasets or real-time applications, while fast simulators lack the fidelity needed for soft materials. Data-driven approaches, though promising, require tens of thousands of physical samples, an unrealistic demand for fragile sensors that cannot withstand extensive testing. Each measurement risks degrading or destroying the sensor, making large-scale data collection impractical. In this work, we focus on a planar, two-dimensional tactile deformation problem as a controlled setting for developing and validating the proposed minimal-data modelling pipeline. To address these challenges, we introduce a data-efficient approach that limits reliance on real sensor measurements without sacrificing accuracy, built on three components: FEM validation on a small real dataset, synthetic data generation, and fast GNN surrogates. 

We demonstrate our method on a planar tactile skin, using camera-tracked tests to achieve sub-0.5 mm accuracy in over 90\% of cases with only a few hours of physical data collection. The validated FEM model produces more than 700 synthetic contact scenarios, spanning diverse positions and indentation depths. GNNs trained on this synthetic data reach a mean absolute error of 0.32mm and perform equally well on real sensor measurements, confirming strong sim-to-real transfer. We show that a small set of real measurements can be used to validate a high-fidelity FEM model, enabling synthetic-data training of a fast GNN surrogate that transfers accurately to a physical soft tactile sensor. Our main contributions are:

\begin{itemize}

\item A data-efficient modelling framework for a planar 2D soft tactile structure that uses only 82 real measurements to validate a high-fidelity FEM model and then generates a larger synthetic dataset for learning, reducing the need for extensive physical testing of fragile tactile structures. 

\item Bidirectional Graph Neural Network surrogate models for tactile deformation prediction and contact inference, trained entirely on FEM-generated data and achieving 0.32 mm mean absolute error with real-time inference. 

\item Experimental sim-to-real validation showing that models trained only on synthetic data maintain sub-0.5 mm accuracy on real measurements, supporting the use of FEM-validated surrogates for design exploration and simulation integration.\end{itemize}

The paper is structured as follows: Section II reviews related work. Section III details the experimental setup, data collection, FEM validation, synthetic data generation, and the GNN architecture and training. Section IV presents experimental results. Section V discusses implications and limitations. Section VI concludes and discusses future work.

\section{Related Work}

\subsection{Soft Material Deformation Modelling}
Accurately predicting how soft materials deform is key for robotics, from manipulation to wearables and sensor design. Finite Element Methods (FEM) offer sub-millimetre accuracy for hyperelastic materials \cite{lin2023comparative}, but their high computational cost makes them impractical for large datasets or real-time use \cite{qin2024modeling}.
Researchers have sought to accelerate soft-material simulation using reduced-order models and real-time FEM. For example, Goury and Duriez \cite{goury2018fast} developed a model-order reduction technique that speeds up FEM for soft robot control while maintaining accuracy. However, these methods still depend on specialised solvers and are costly for generating large datasets.
Engines such as MuJoCo \cite{todorov2012mujoco}, Bullet \cite{coumans2015bullet}, and Isaac Gym \cite{makoviychuk2021isaac}, among others, enable real-time simulation by simplifying contact handling and using rigid-body models \cite{zhang2026embodied}. Even with recent updates to handle deformables, these engines often rely on mass-spring systems or linearised elasticity, which struggle to model large deformations and the hyperelastic behaviour of soft materials \cite{duriez2016soft}.
Vision-based tactile sensors like TacTip \cite{ward2018tactip}, Gelsight \cite{yuan2017gelsight}, and DIGIT \cite{lambeta2020digit} highlight these simulation hurdles. SimTac~\cite{zhang2026simtac} achieves real-time tactile simulation using the Material Point Method (MPM), but even these advanced frameworks still require substantial calibration data to estimate material parameters. Simplified variants, such as tactile contact lenses \cite{kikuuwe2004tactile}, provide a tractable baseline for methodology development while reducing geometric complexity. Our work uses a planar 2D tactile structure, inspired by contact lens designs, as a controlled test case for validating the proposed minimal-data approach before extending the methodology to more complex three-dimensional sensors.

\subsection{Data-Efficient Learning for Physical Systems}
Machine learning for physical systems faces a trade-off: simulators give unlimited synthetic data but are slow, while real experiments provide ground truth but are time-consuming and risky for delicate setups.
Domain randomization~\cite{peng2018sim, tobin2017domain} uses varied simulation parameters to boost real-world robustness, but it requires thousands of simulated samples and can’t fix core simulator flaws. System identification~\cite{thuruthel2018model} matches simulators to real systems but needs hundreds or thousands of real measurements. Residual learning~\cite{ajay2018augmenting} tries to correct simulator errors with neural networks, but still depends on large real datasets.
Pure data-driven methods~\cite{sanchez2018graph, matas2018sim} model deformable dynamics from experiments and work well for soft robots and object manipulation. But these approaches still require tens of thousands of samples to generalise, which is not feasible for fragile tactile sensors.
Recent advances, such as RoboCraft~\cite{shi2024robocraft}, show how learning and simulation can work together for deformable modelling, but still require large training datasets.

\subsection{Graph Neural Networks for Physics Simulation}
Graph Neural Networks (GNNs) are well-suited for capturing relationships and interactions in dynamical systems. Graph Networks~\cite{battaglia2018relational} and the Graph Network Simulator~\cite{sanchez2020learning} demonstrated that GNNs can learn particle-based interactions. MeshGraphNets~\cite{pfaff2020learning} extended this to mesh-based simulations, enabling the prediction of deformable objects and fluid dynamics up to 1000× faster than traditional solvers.
Recently, GNN surrogate models have been used to speed up FEM in robotics; for example, this paper~\cite{duret2026real} uses GNNs for deformable grasping, approximating FEM deformation and stress fields with significant speed gains. However, these models are validated only against FEM simulations, not real hardware, leaving the sim-to-real gap unaddressed, a critical limitation for deployment on physical systems.

 \medskip Our work stands out from previous approaches in three ways. First, we show that just 82 selected measurements are sufficient to validate high-fidelity FEM, thereby enabling scalable synthetic data generation. Second, unlike prior GNN surrogates, we provide real-world validation confirming that models trained entirely on synthetic data remain accurate on physical sensors, while using an order of magnitude fewer real data than neural network methods, which usually require 1000s of samples. Third, we enable real-time deformation visualisation in simulation environments, something earlier GNN-based physics learning approaches lack, as they focus only on offline accuracy. Together, these advances make practical virtual prototyping and design exploration possible without the need for extensive sensor fabrication or testing.

\section{Methodology}

 \begin{figure*}[th]
      \centering
       \vspace{0.1cm}
      \includegraphics[height= 12.5cm]{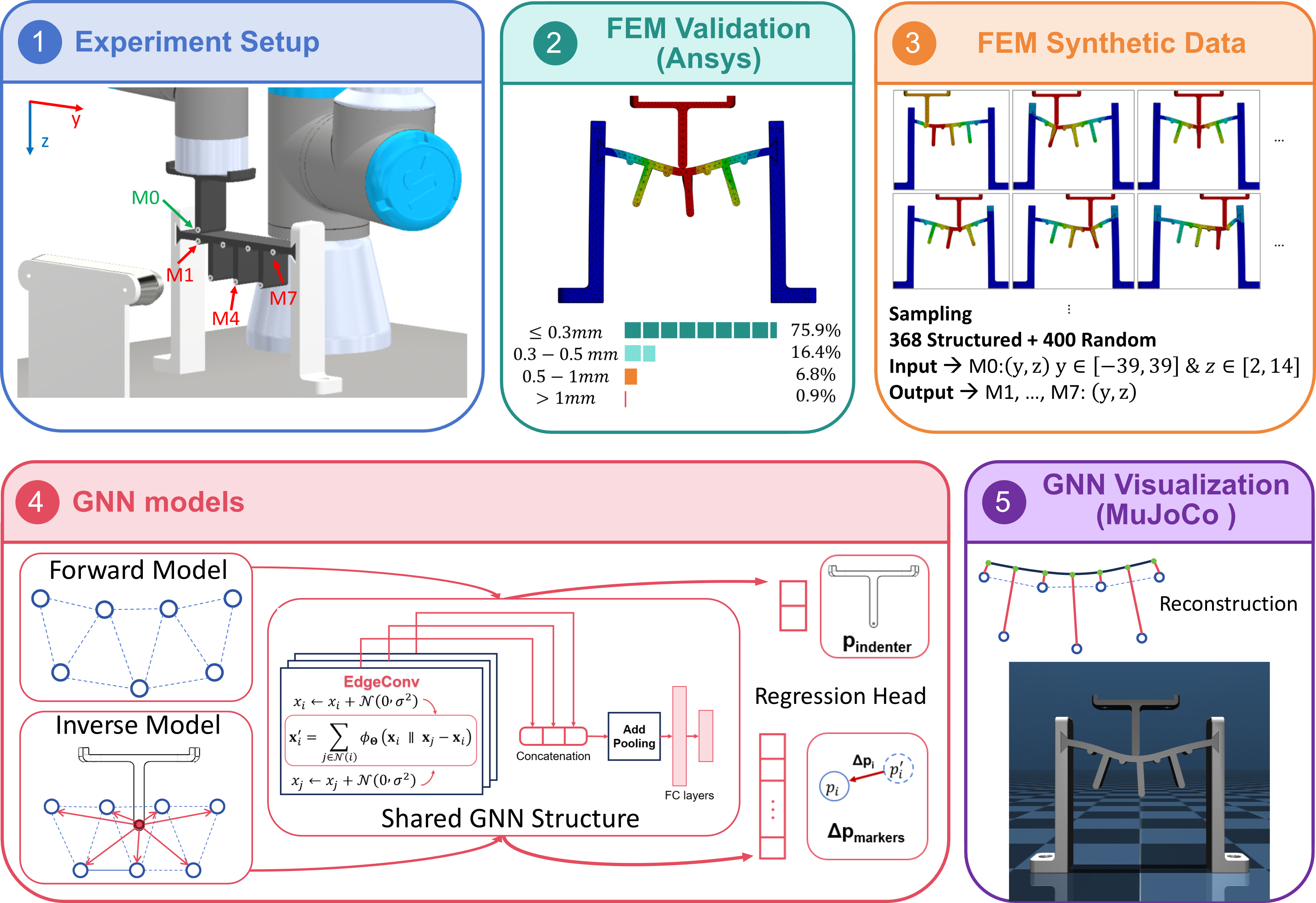}
      \caption{Overview of the proposed framework for tactile deformation modelling and real-time visualisation. (1) Experimental setup of the tactile sensing system, consisting of a compliant TPU structure constrained between rigid supports and indented at different spatial locations (y,z). Marker M0 denotes the indenter marker, while markers M1–M7 denote the tracked pin markers. (2) Finite element model (ANSYS) validation against experimental measurements to verify deformation accuracy. (3) Generation of a synthetic dataset through 368 structured and 400 random parametric FEM simulations by varying indenter positions, extracting the y and z coordinates of the seven markers for each case. (4) Training of Graph Neural Network (GNN) models using the FEM-generated dataset to learn the mapping between indenter position and marker displacements. (5) Real-time visualisation of predicted structural deformation in MuJoCo based on GNN inference.
      }
      \label{fig1}
   \end{figure*}

Fig. \ref{fig1} summarises the proposed pipeline: experimental measurements are used to validate the ANSYS FEM model, the validated model generates structured and random synthetic contact configurations, and the resulting data are used to train bidirectional GNN surrogates for real-time deformation visualisation in MuJoCo.

\subsection{System Overview}

The soft morphological skin used in this study is a tactile sensing system comprising a planar, compliant structure, 10 cm long, fabricated from thermoplastic polyurethane (TPU 95A) that deforms under external contact. This planar morphology was chosen to define a controlled 2D deformation problem, facilitating FEM validation and systematic data collection while retaining contact behaviours relevant to robotic manipulation. A series of pins of two alternating lengths extrudes from the structure. The pins amplify and spatially distribute the deformation induced by contact. The compliant structure is mechanically constrained between two rigid supports manufactured in polylactic acid (PLA), which provide fixed boundary conditions during interaction.

A rigid cylindrically-tipped indenter mounted on a UR5 robot arm is used to apply controlled contact to the structure. During interaction, the indenter moves along the in-plane direction (y-axis in Fig.~\ref{fig:stacked}) and presses against the TPU structure, producing deformations primarily within the two-dimensional yz plane defined by the sensor geometry. Displacements in the out-of-plane direction (x-axis) are negligible and are therefore not considered in this work.

The structure's deformation is measured at discrete locations corresponding to physical markers placed on the pins. These marker displacements constitute the sensing system's observable output and are used for both simulation validation and data-driven modeling.

\subsection{Real Data Collection for Validation}
A computer vision system is mounted at the front of the TPU structure, capturing and processing images to generate real-world data. The camera was pre-calibrated using OpenCV's calibration library to correct for lens distortion, such that pixel distances in the image are proportional to physical distances within a given depth plane. White annular rings around the pin and indenter markers serve to isolate the central black dot, the detection target, from the background during binary thresholding. Since all markers lie in the same plane, a single pixel-to-millimetre scaling factor was pre-computed for that plane and applied uniformly to all captured images.

Marker detection is performed using OpenCV's SimpleBlobDetector, with filters on colour, area, circularity, and inertia ratio applied to reject spurious detections. Once the indenter marker and the seven pin markers are identified, each marker is assigned a unique index corresponding to the node and edge indices of the pre-defined graph structure, ensuring that the neighbourhood relationships between markers remain consistent across all deformation states (as illustrated in Fig. \ref{fig:stacked}).

The dataset was collected by capturing images at 82 fixed contact configurations, obtained by translating the indenter along the in-plane direction (y-axis) and displacing it along the normal direction (z-axis) from $2$ to $14$~mm in $2$~mm increments, as summarized in Table~\ref{tab:data_collection}. Due to the mechanical constraints of the PLA lateral supports, the achievable indentation depth is shallower near the two ends of the soft morphological skin than at the centre. The data demonstrated good reproducibility and stability, with an average positional mean error of $0.063$~mm and an average standard deviation of $0.066$~mm across six repeated trials.


\begin{table}[htbp]
\centering
\small
\setlength{\tabcolsep}{6pt}
\renewcommand{\arraystretch}{1.15}

\caption{Indenter position ranges and step sizes for data collection. The origin of the coordinate system is defined at the centre of the skin surface.}
\label{tab:data_collection}

\begin{tabular}{lcccc}
\toprule
 & \multicolumn{3}{c}{Range (mm)} & Step \\

\midrule
Y (mm) & [-39, -30] & (-30, 30) & [30, 39] & 3 \\
Z (mm) & [2, 6]     & [2, 14]   & [2, 6]   & 2 \\

\bottomrule
\end{tabular}
\end{table}

\subsection{High-Fidelity FEM Simulation}
A finite element model was developed in ANSYS to simulate how the tactile structure responds to contact, and its predictions were validated against experimental measurements. The skin was defined as a deformable body with a hyperelastic material model, while the indenter and lateral supports were treated as rigid bodies, with boundary conditions reflecting the physical PLA constraints.

Thermoplastic polyurethane (TPU) exhibits highly nonlinear stress-strain behavior under large deformations, requiring hyperelastic constitutive models. The Ogden model was selected because it accurately captures the strain-stiffening behaviour characteristic of 3D printed TPU parts under large deformations \cite{martin2025anisotropic}. The physical skin was fabricated from 3D JAKE TPU A95 Black filament. Since manufacturer-provided material data were unavailable, Ogden parameters were adopted from the experimental characterisation of NinjaFlex Cheetah 95A TPU~\cite{smith2024mechanical}, a comparable material with the same Shore hardness. 

Contact between the indenter and the skin was modelled as frictional, with a coefficient of 0.5. Although this is higher than typical TPU-PLA friction values (0.2–0.3), it prevented unrealistic sliding during indentation and more accurately reproduced observed deformation. The increased coefficient likely reflects surface roughness from 3D printing or partial adhesive effects.

 The TPU structure was discretised using second-order tetrahedral elements (SOLID187). The model was meshed using ANSYS automatic meshing at a resolution setting of 7, and large-deformation effects were enabled to account for geometric nonlinearity.
Each indentation scenario was simulated individually under quasi-static conditions, with the indenter placed at various in-plane positions and pressed to depths from 2 to 14 mm in 2 mm steps as in the experimental data. We extracted nodal displacements at the locations of physical markers, considering only the in-plane (y and z) components, as out-of-plane motion was negligible. These displacement values served as high-fidelity reference data for GNN training.

\subsection{Synthetic Data Set Generation}
Once validated against 82 physical measurements, the FEM model was used to generate an expanded synthetic dataset for GNN training. The complete dataset comprises 768 contact configurations, obtained using two complementary sampling strategies:
\begin{itemize}
    \item \textbf{Structured sampling (368 configurations).} Contact configurations follow a regular grid pattern with systematic variation in position and depth. The 82 physically validated configurations and the 286 additional grid-based configurations maintain systematic sampling $(z \in \{2, 4, 6, 8, 10, 12, 14\}\ \mathrm{mm})$ with regular $y$-axis spacing
    \item \textbf{Random sampling (400 configurations).}  To ensure the GNN generalises beyond regular grid points, we generated 400 additional configurations using continuous randomisation. Y-positions were uniformly sampled from [-39, 39] mm and Z-depths from [2, 14] mm, both as continuous values.
\end{itemize}

This hybrid sampling strategy combines the interpretability of structured data with the diversity of random sampling, ensuring robust GNN training. The random samples expose the network to contact configurations between grid points, improving interpolation capability and generalisation.
For each configuration, ANSYS simulations were performed following the protocol in Section III-C, and nodal displacements at seven marker locations were extracted as labels.
This approach demonstrates the data efficiency of our framework: only 82 physical measurements validate the FEM model, which then generates nearly 9× more training data (768 total) without additional sensor wear.

\subsection{Graph Neural Network Models}

The proposed bidirectional GNN models comprise two branches which differ in their input features and prediction targets. The forward model performs contact inference: the node feature matrix $\mathbf{x}_{\text{fwd}} \in \mathbb{R}^{7 \times 2}$
includes the absolute positions of the seven pin markers $\mathbf{p}^i_{\text{mkr}}$, and the model predicts the indenter position ${\mathbf{p}}_{\text{ind}} \in \mathbb{R}^{1 \times 2}$. The inverse model performs tactile deformation prediction: the node feature matrix $\mathbf{x}_{\text{inv}} \in \mathbb{R}^{7 \times 2}$ includes the relative displacements between the indenter marker and each pin marker, defined as $\mathbf{p}_{\text{ind}} - \mathbf{p}^i_{\text{mkr}}$, and the model predicts the full set of pin marker displacements $\Delta{\mathbf{p}}_{\text{mkr}} \in \mathbb{R}^{7 \times 2}$.

Both models share an identical graph topology which is defined solely over the pin markers according to a predetermined neighborhood relation, yielding 11 undirected edges. Under the standard bidirectional edge convention adopted in message-passing networks in GNN, each undirected edge is represented as two directed edges, resulting in an edge index tensor $\mathbf{E} \in \mathbb{Z}^{2 \times 22}$.

The forward and inverse models use the same GNN structure and hyperparameters but differ in the input dimension of the first graph convolution layer and the output dimension of the final fully connected layer, as dictated by their respective tasks. They were trained individually, so they do not share weights. Both models are trained with mean squared error (MSE) as the loss function: for the forward model, MSE is computed over the predicted indenter position, whereas for the inverse model, it is computed over the predicted displacements of the seven markers. Each model was trained on 768 synthetic samples, split into a training set (70\%) and a validation set (30\%). An additional 400 real-world samples, collected under contact configurations different from those used in the synthetic dataset, were held out as an independent real-world validation set.

The model employs the EdgeConv operation proposed in DGCNN \cite{wang2019dynamic}, which aggregates both the nodal displacement values and the pairwise differences between a node and its neighbours to construct a local graph. The difference is that we do not use dynamically generated edge indices; instead, we fix the edges for each node. The architecture comprises a graph convolution module consisting of several EdgeConv layers; an aggregation layer that concatenates the outputs of all EdgeConv layers to obtain a global receptive field; a global pooling layer for compressing the feature representation; and a fully connected module that decodes the high-dimensional feature. Since the GNN is trained exclusively on data synthesised in simulation, Gaussian noise is injected into the marker positions during training to improve sim-to-real generalisation. Specifically, the perturbed position is given by $\tilde{\mathbf{x}} = \mathbf{x} + \boldsymbol{\varepsilon}$, $\boldsymbol{\varepsilon} \sim \mathcal{N}(\mathbf{0}, \sigma^{2}\mathbf{I})$, where $\sigma$ controls the noise magnitude. Hyperparameter search was conducted to optimise model performance; the selected configuration is reported in Table~\ref{tab:hyperparameter}. 

\begin{table}[h]
\centering
\renewcommand{\arraystretch}{1.2} 
\begin{tabularx}{\columnwidth}{X l}
\toprule
\textbf{Parameter} & \textbf{Value} \\ \midrule
Convolutional hidden layers & [64, 64, 128, 256] \\
Concat hidden layers        & 1024 \\
FC hidden layers            & [512, 256] \\
Hidden-layer activation     & LeakyReLU \\
Dropout coefficient         & 0.2 \\
Noise magnitude             & 0.5 \\
Batch size                  & 128 \\ \bottomrule
\end{tabularx}
\caption{GNN hyperparameter configurations.}
\label{tab:hyperparameter}
\end{table}

\section{Results}
This section presents results across three stages of the pipeline for the planar 2D tactile deformation task: FEM validation against real measurements, GNN training and validation on synthetic data, and sim-to-real transfer to the physical sensor. Table~\ref{tab:summary} summarises the main quantitative outcomes across all three stages, with detailed analysis in the subsections below. The GNN achieves inference in approximately 4ms per sample ($\sim$0.004s), compared to $\sim$15 minutes for a single 
ANSYS FEM simulation, representing a speedup of over 200,000$\times$. The complete MuJoCo visualisation loop, including GNN inference and rendering, runs at 10ms per frame on a consumer-grade CPU (AMD Ryzen 5800H), confirming real-time suitability.
\begin{table}[ht]
\centering
\caption{Summary of main quantitative results.}
\label{tab:summary}
\resizebox{\columnwidth}{!}{
\begin{tabular}{lcccc}
\toprule
& FEM Real Val. & GNN Synthetic Val. & GNN Real Val.  \\
\midrule
Y & 90.6\% $\leq$0.5 mm & MAE 0.43 mm & MAE 0.80 mm &  \\
Z & 93.9\% $\leq$0.5 mm & MAE 0.10 mm & MAE 0.41 mm & \\
\bottomrule
\end{tabular}
}
\end{table}
\noindent 
\subsection{FEM Validation Results on Real Measurements}
To assess model accuracy, we compared simulated nodal displacements with camera-tracked measurements from the physical sensor under the same loading conditions. Table~\ref{tab:fem_accuracy} shows the percentage of samples within different absolute error thresholds for Y, Z, and combined (Y+Z) displacements.

The FEM results closely match experimental data: 82.2\% of Y-direction and 90.6\% of Z-direction samples have errors below 0.3 mm. With a 0.5 mm threshold, appropriate for the sensor’s millimeter scale features, over 90\% of Y and 93.9\% of Z samples are within this range. For combined displacement, 75.9\% of samples are below 0.3 mm, and 92.3\% are below 0.5 mm. Errors greater than 1 mm are rare, seen in only 1.9\% of Y samples and none of the Z samples.

These results confirm that the ANSYS FEM model reliably predicts both the magnitude and spatial distribution of deformation under contact. With this validated model, we can confidently generate synthetic training data without further physical testing that could damage the sensor.

\begin{table}[ht]
\centering
\small
\setlength{\tabcolsep}{4pt}
\renewcommand{\arraystretch}{1.15}

\resizebox{\columnwidth}{!}{%
\begin{tabular}{llcccc c}
\toprule

\multirow{2}{*}{\textbf{}} & \multirow{2}{*}{\textbf{}} 
& \multicolumn{4}{c}{Error range (\%)} 
& \multirow{2}{*}{$(Y{+}Z)\leq 0.5$} \\
\cmidrule(lr){3-6}

Error thresholds (mm) & & $E\leq0.3$ & $E\leq0.5$ & $0.5<E\leq1$ & $E>1$ & \\

\midrule

\multirow{2}{*}{M1} & y & 98.8 & \textcolor{green!60!black}{100} & 0 & 0 &  \multirow{2}{*}{99.4} \\ & z & 69.5 & \textcolor{green!60!black}{98.8} & \textcolor{orange!85!black}{1.2} & 0 & \\

\addlinespace[2pt]

\multirow{2}{*}{M2} & y & 42.7 & \textcolor{green!60!black}{67.1} & \textcolor{orange!85!black}{26.8} & \textcolor{red!75!black}{6.1} & \multirow{2}{*}{79.9} \\
                    & z & 68.3 & \textcolor{green!60!black}{92.7} & \textcolor{orange!85!black}{7.3} & 0 & \\

\addlinespace[2pt]

\multirow{2}{*}{M3} & y & 100 & \textcolor{green!60!black}{100} & 0 & 0 & \multirow{2}{*}{92.1} \\
                    & z & 54.9 & \textcolor{green!60!black}{84.2} & \textcolor{orange!85!black}{15.9} & 0 & \\

\addlinespace[2pt]

\multirow{2}{*}{M4} & y & 39.0 & \textcolor{green!60!black}{67.1} & \textcolor{orange!85!black}{25.6} & 0 & \multirow{2}{*}{74.4} \\
                    & z & 41.5 & \textcolor{green!60!black}{81.7} & \textcolor{orange!85!black}{18.3} & 0 & \\

\addlinespace[2pt]

\multirow{2}{*}{M5} & y & 100 & \textcolor{green!60!black}{100} & 0 & 0 & \multirow{2}{*}{100} \\
                    & z & 56.1 & \textcolor{green!60!black}{100} & 0 & 0 & \\

\addlinespace[2pt]

\multirow{2}{*}{M6} & y & 96.3 & \textcolor{green!60!black}{100} & 0 & 0 & \multirow{2}{*}{100} \\
                    & z & 96.3 & \textcolor{green!60!black}{100} & 0 & 0 & \\

\addlinespace[2pt]

\multirow{2}{*}{M7} & y & 98.8 & \textcolor{green!60!black}{100} & 0 & 0 & \multirow{2}{*}{100} \\
                    & z & 100  & \textcolor{green!60!black}{100} & 0 & 0 & \\

\midrule

\multirow{2}{*}{Total} & y & 82.2 & \textcolor{green!60!black}{90.6} & \textcolor{orange!85!black}{7.5} & \textcolor{red!75!black}{1.9} & \multirow{2}{*}{92.3} \\
                       & z & 69.5 & \textcolor{green!60!black}{93.9} & \textcolor{orange!85!black}{6.1} & 0 & \\

\bottomrule
\end{tabular}
}
\caption{Validation of ANSYS FEM Against Physical Measurements. Cumulative percentages of samples achieving specified displacement error thresholds in mm for individual markers (M1-M7) and aggregated results (Total). The final column reports joint Y+Z accuracy below 0.5mm threshold. N=82 validation configurations.}
\label{tab:fem_accuracy}
\end{table}

\subsection{GNN Validation on Real Sensor Measurements}

The forward model, which predicts indenter position from marker observations, achieves an average synthetic-validation MAE of 0.43 mm in Y and 0.10 mm in Z, and an average real-validation MAE of 0.80 mm in Y and 0.41 mm in Z. The generalisation gap is 0.37 mm in Y and 0.31 mm in Z, corresponding to 0.4\% and 2.5\% of the indenter's designed range of motion, respectively. Scatter plots of the synthetic-validation and real-validation predictions for one representative fold are shown in Fig. \ref{fig:scatter}; larger errors are concentrated at the extremes of the output range.

\begin{table}[htbp]
    \centering
    \caption{MAE results for the displacement $\Delta{\mathbf{p}}_{\text{mkr}}$ (mm) of each marker along the Y and Z axes.}
    \label{tab:inverse_mae_markers}
    \resizebox{\columnwidth}{!}{%
    \renewcommand{\arraystretch}{1.1}
    \begin{tabular}{llccccccccc}
        \toprule
        \multicolumn{2}{c}{} & \multicolumn{7}{c}{Pin Marker Number (\#)} & \\
        \cmidrule(lr){3-9}
        & & M1 & M2 & M3 & M4 & M5 & M6 & M7 & Avg. \\
        \midrule
        \multirow{2}{*}{Y}
            & Syn. & 0.0608 & 0.1780 & 0.0625 & 0.1525 & 0.0611 & 0.1523 & 0.0579 & 0.1036 \\
            & Real       & 0.0607 & 0.4916 & 0.0684 & 0.4245 & 0.0519 & 0.2006 & 0.0601 & 0.1940 \\
        \midrule
        \multirow{2}{*}{Z}
            & Syn. & 0.1348 & 0.1378 & 0.1387 & 0.1443 & 0.1297 & 0.1456 & 0.1513 & 0.1403 \\
            & Real       & 0.1743 & 0.2827 & 0.4612 & 0.5458 & 0.4882 & 0.2994 & 0.1732 & 0.3464 \\
        \bottomrule
    \end{tabular}%
    }
\end{table}

Table \ref{tab:inverse_mae_markers} reports the average per-marker MAE in the Y and Z directions across the five folds of the inverse model. In the Y-direction, the four surface-mounted pin markers (\#1, 3, 5, 7) show little change between the mean error on the synthetic-validation set and on the real-validation set, with both remaining around 0.06 mm. By contrast, the three extended pin markers (\#2, 4, 6) exhibit a more pronounced increase in test error, as their greater length amplifies the effect of local surface slope variations. In the Z-direction, the three central pin markers (\#3, 4, 5) exhibit a notably larger test error, which is attributable to the greater indentation depth at the centre of the structure relative to its ends, as imposed by the experimental design. Scatter plots of the synthetic-validation and real-validation predictions for one representative fold are shown in Fig. \ref{fig:scatter}, with all seven markers plotted jointly. The Y- and Z-directions ranges correspond to the minimum and maximum displacements in each respective direction.

    
    

\begin{figure*}[t]
    \centering
    \setlength{\tabcolsep}{6pt}
    \renewcommand{\arraystretch}{1.2}

    \newcommand{\rowtitle}[1]{%
        \raisebox{1.2em}{%
            \parbox{0.10\textwidth}{\centering\textbf{#1}}%
        }%
    }

    \centering
    \makebox[\textwidth][l]{%
    \begin{tabular}{
        @{}
        >{\raggedright\arraybackslash}m{0.07\textwidth}
        >{\centering\arraybackslash}m{0.43\textwidth}
        >{\centering\arraybackslash}m{0.43\textwidth}
        @{}
    }
        & \textbf{Synthetic data} & \textbf{Real data} \\[0.5em]

        \rowtitle{Forward model} &
        \includegraphics[width=\linewidth]{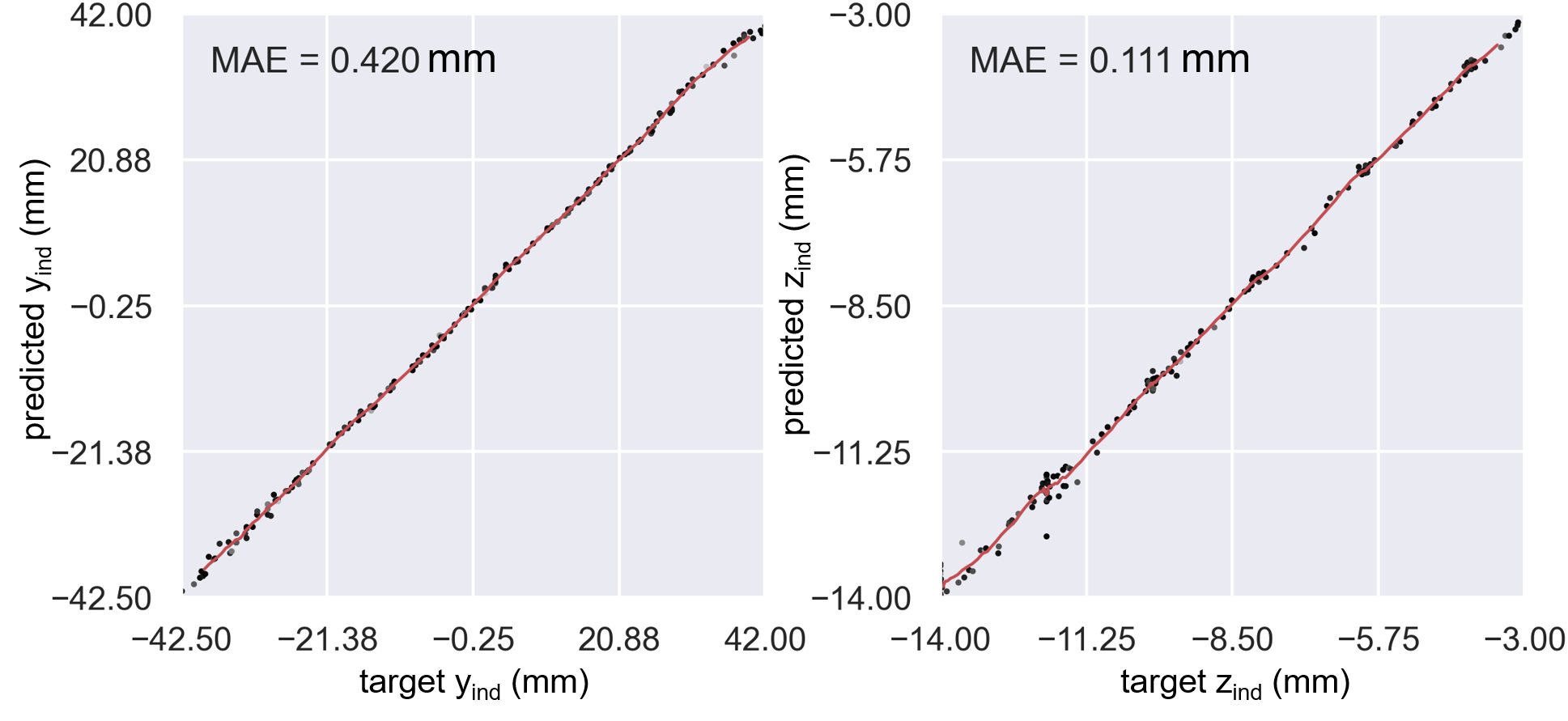}
        &
        \includegraphics[width=\linewidth]{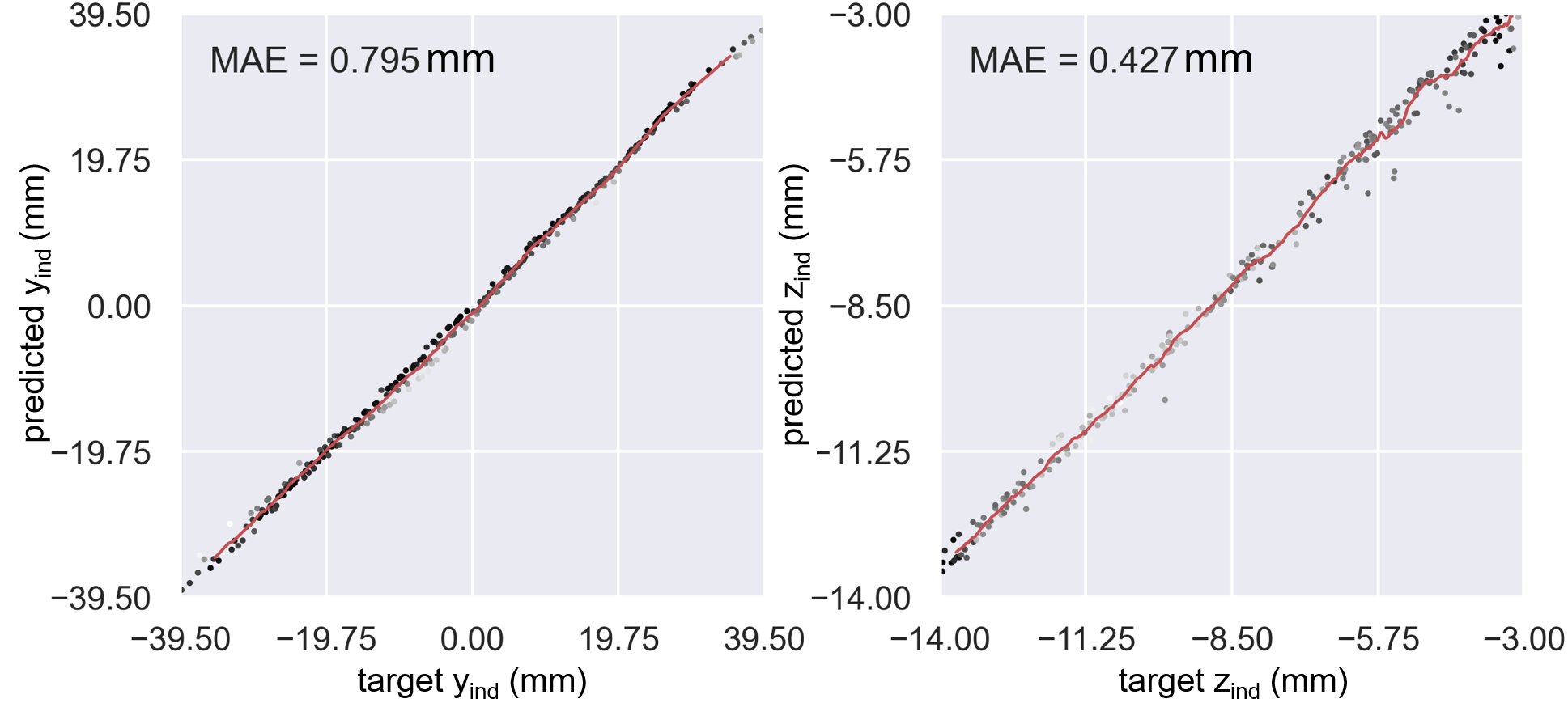}
        \\[-0.2em]

        &
        \multicolumn{2}{>{\centering\arraybackslash}p{0.82\textwidth}}{
            \small
            Indenter marker position predicted from pin marker positions.
        }
        \\[1.2em]

        \rowtitle{Inverse model} &
        \includegraphics[width=\linewidth]{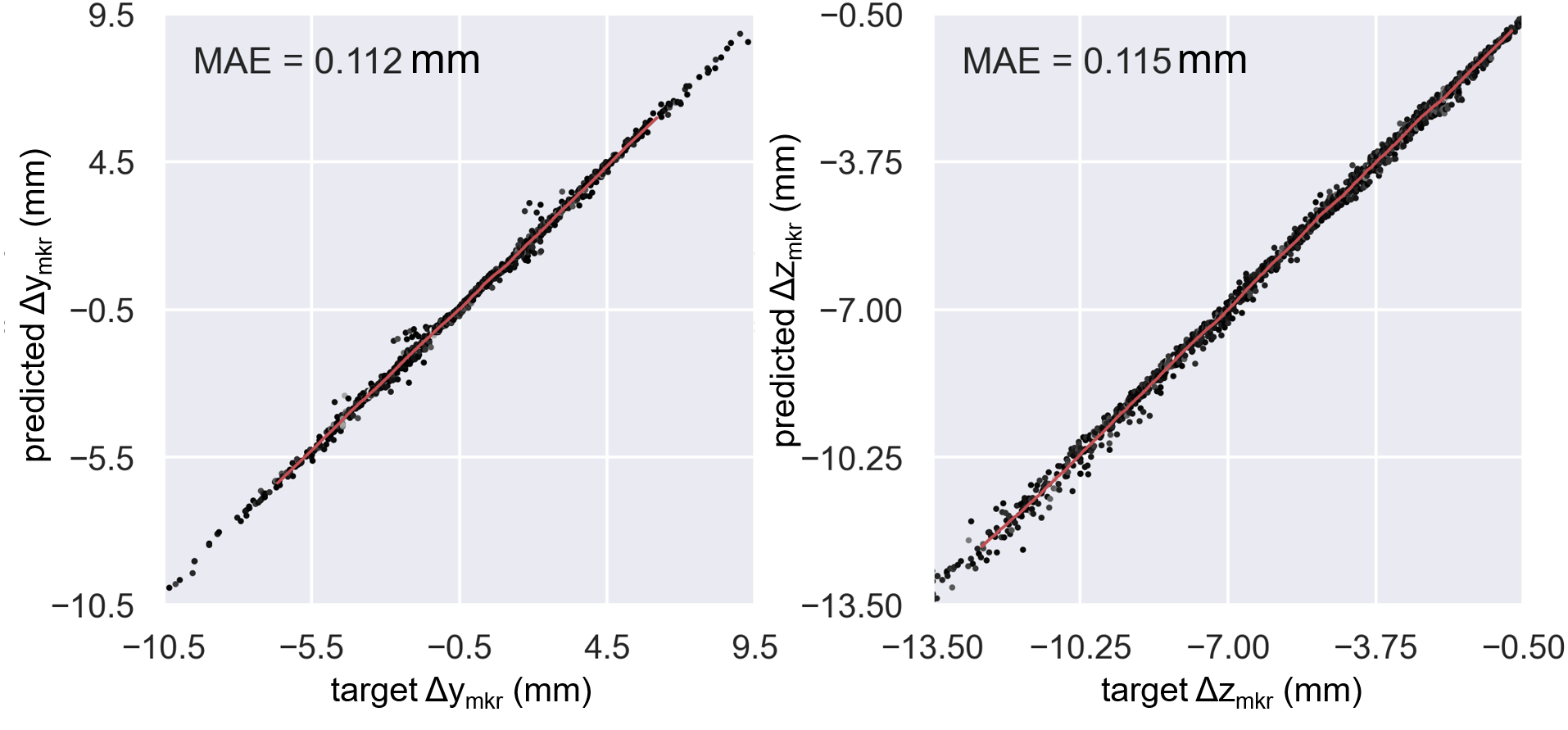}
        &
        \includegraphics[width=\linewidth]{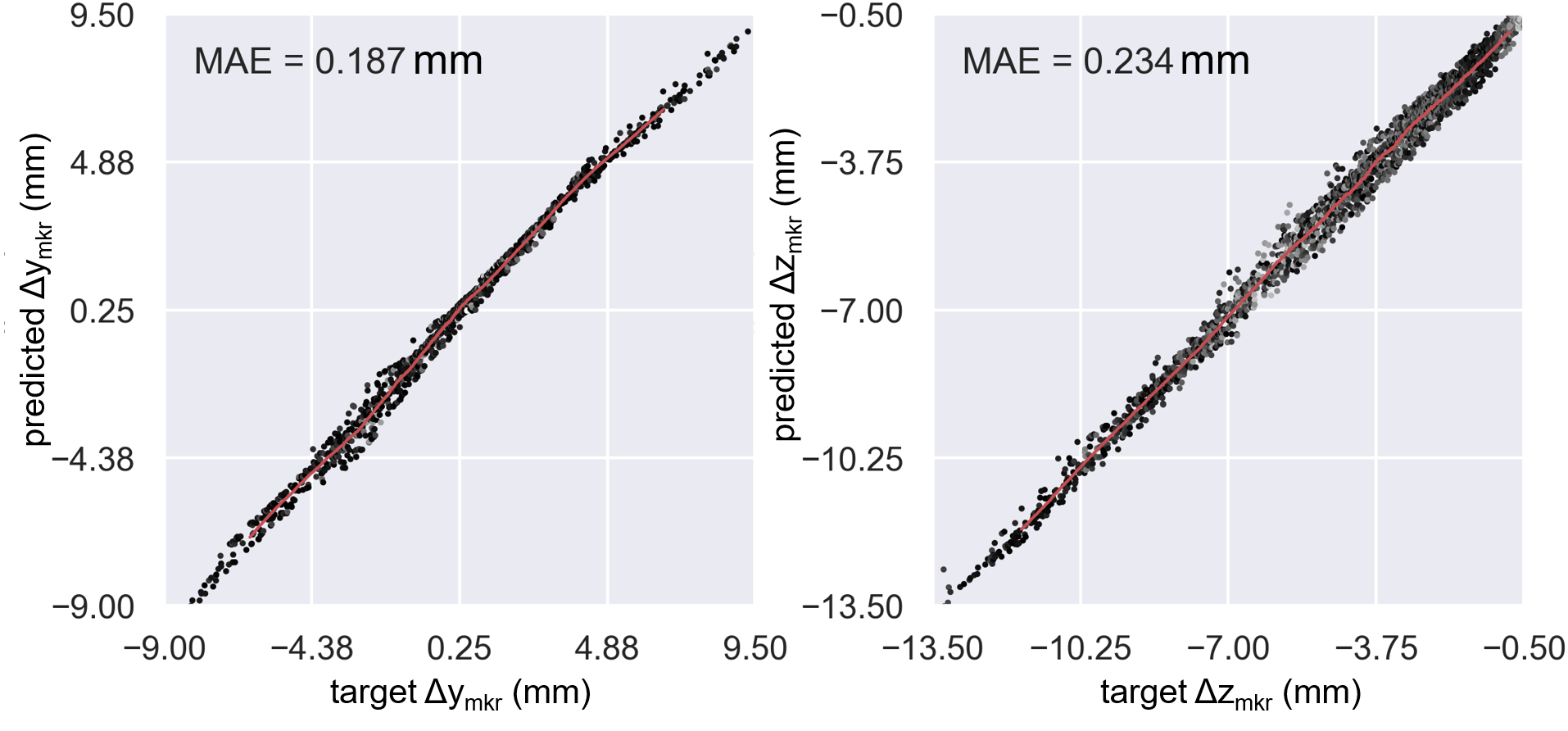}
        \\[-0.2em]

        &
        \multicolumn{2}{>{\centering\arraybackslash}p{0.82\textwidth}}{
            \small
            Pin marker displacements predicted from relative indenter--pin marker positions.
        }
    \end{tabular}
    }
    \caption{Scatter plots for the forward and inverse models on synthetic and real data. The forward model predicts the absolute position of the indenter marker from the absolute positions of the pin markers. 
    The inverse model predicts the displacements of all seven pin markers from the relative positions between the indenter marker and the pin markers.}
    \label{fig:scatter}
\end{figure*}

\begin{figure}[htbp]
    \centering
    \begin{subfigure}[b]{0.46\columnwidth}
        \centering
        \includegraphics[height=3cm]{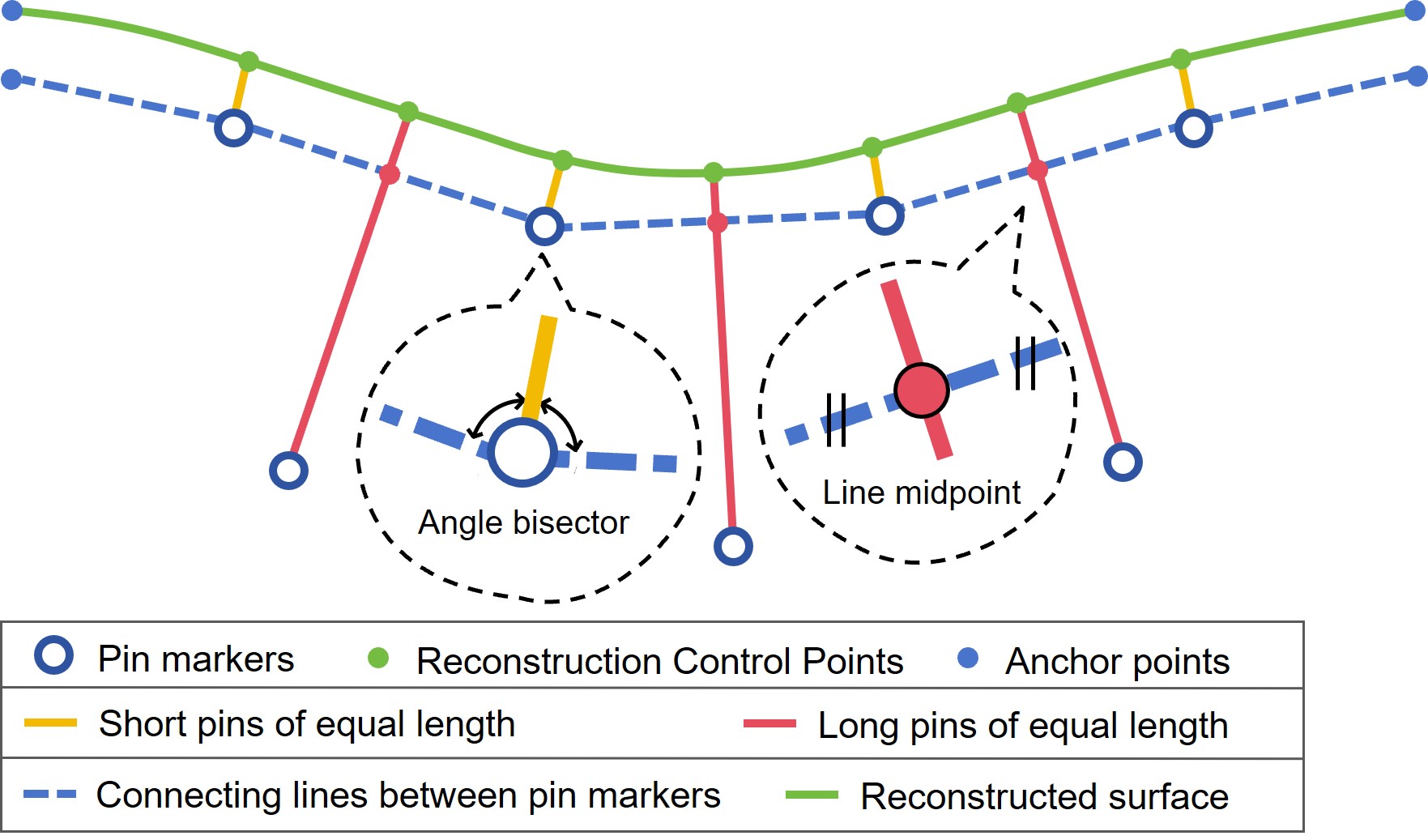}
        \caption{}
        \label{fig:reconstruction-1}
    \end{subfigure}
    \hfill
    \begin{subfigure}[b]{0.46\columnwidth}
        \centering
        \includegraphics[height=3cm]{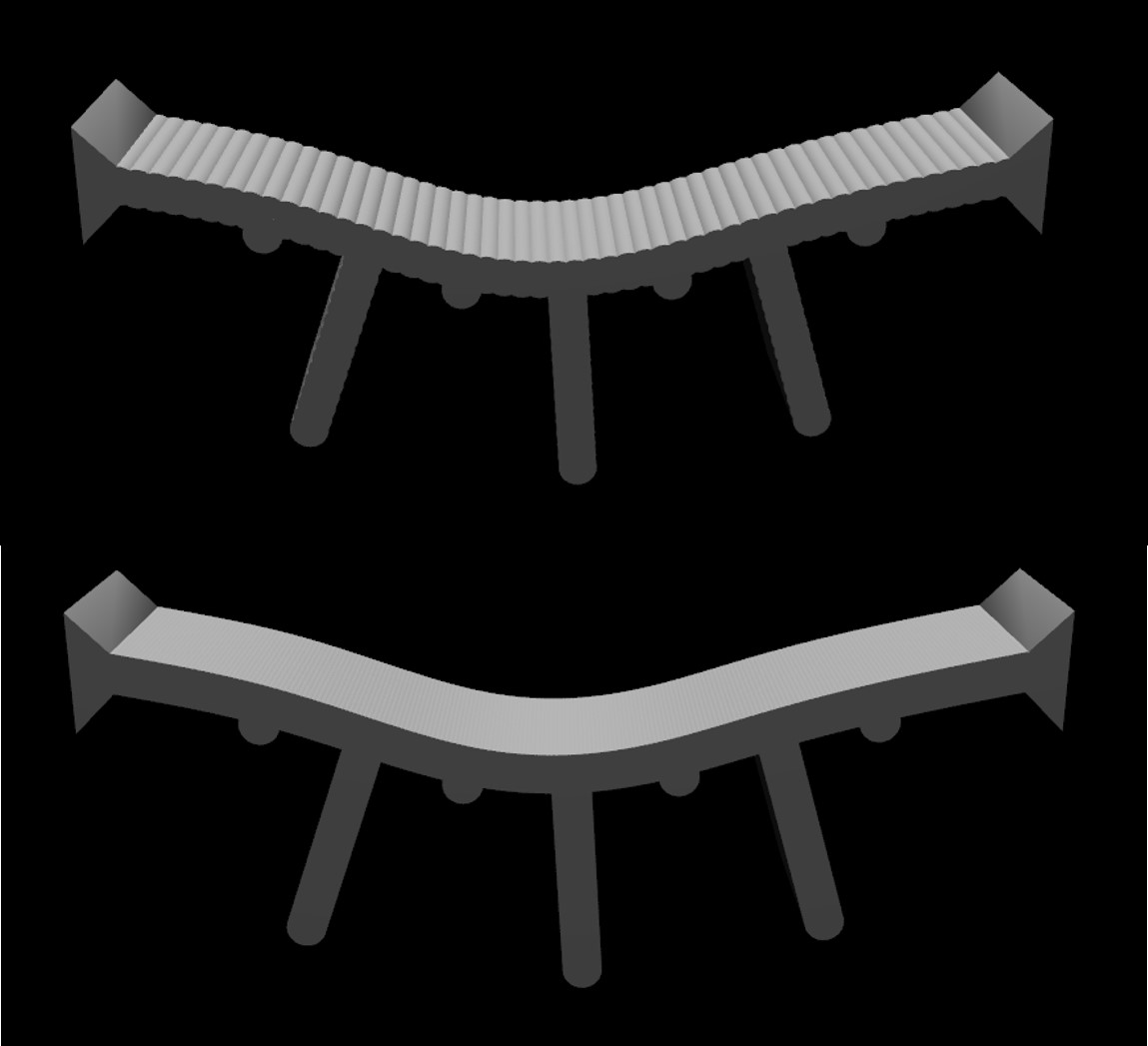}
        \caption{}
        \label{fig:reconstruction-2}
    \end{subfigure}
    \caption{(a): Several geometric constraints, including the fixed length of the pins, angle bisector and line midpoint, define the control points on the surface of the structure. These control points are used to generate the cubic spline equations that fit the surface. (b): A number of standard cylinders arranged at equal intervals used for the visualization in MuJoCo. The number of cylinders determines the level of visual detail: low (top) and high (bottom).}
    \label{fig:reconstruction}
\end{figure}

\subsection{Real-Time Visualization Framework}

The low inference latency of the GNN enables its integration into robot simulation software for real-time visualisation of TPU structure deformation. From the seven predicted pin marker displacements, seven corresponding surface anchor points are derived via geometric constraints; these serve as control points for a cubic spline that reconstructs the deformed surface profile of the structure (Fig. \ref{fig:reconstruction}).

MuJoCo is used as the simulation environment. Two virtual sites, placed at the origin of the TPU structure and at the indenter contact point respectively, are used to compute the relative position between the indenter and the structure, which serves as the input to the GNN. Once the surface curve is calculated, a series of cylinders are uniformly distributed along the surface spline and along each pin segment to approximate the structure geometry, while the indenter and the lateral supports of the TPU structure are represented by pre-imported CAD meshes. Collision detection between the cylinders comprising the TPU structure and the indenter is disabled, while collisions with all other bodies are retained. Although this pipeline does not interact with the MuJoCo physics engine, it produces a visually realistic real-time rendering of the deformable structure. Running on CPU, GNN inference requires approximately 4 ms per sample, while the complete visualisation loop, including inference and rendering, runs at approximately 10 ms per frame.

\section{Discussions}
\subsection{Data Efficiency and Validation Strategy}
We show that, for a planar 2D soft tactile deformation problem, a small set of real measurements can be used to validate a high-fidelity FEM model, enabling synthetic-data training of a fast GNN surrogate that transfers accurately to a physical soft tactile sensor. This result highlights a practical route to data-efficient modelling for fragile soft tactile systems. With only 82 physical measurements, we validated the ANSYS FEM model as an accurate reference model for the tested contact conditions, with over 90\% of samples falling within 0.5 mm error and generated 768 diverse training samples. Data efficiency is particularly important for soft tactile sensors because repeated physical testing can accelerate material wear and increase prototyping cost. This approach offers a practical alternative to purely data-driven pipelines by using real measurements for model validation rather than large-scale training data collection. This is especially valuable because purely data-driven approaches often require substantially larger real datasets for robust generalisation.

\subsection{Sim-to-Real Transfer Performance}
The GNN models show good sim-to-real transfer, with models trained entirely on synthetic FEM data retaining sub-millimetre accuracy on real sensor measurements. The forward model achieves an average test error of 0.80 mm in the Y-direction and 0.41 mm in the Z-direction on real sensor measurements. The generalization gap, 0.37 mm (Y) and 0.31 mm (Z), is less than 3\% of the sensor's operational range. These results suggest that the validated FEM captures the dominant quasi-static deformation behaviour of the sensor under the tested contact conditions.

The inverse model shows varying performance across markers. 
Surface-mounted pins (\#1, 3, 5, 7) consistently achieve low errors across the entire workspace, as their predominantly translational motion is captured equally well by FEM and camera tracking. Among extended pins, \#6 achieves comparable accuracy, while \#2 and \#4 show a slightly larger sim-to-real gap. One possible explanation is a planar imaging effect: under larger indentations, extended pins may undergo small rotations whose out-of-plane components are represented in the FEM but projected differently onto the camera image plane. Since the discrepancies remain below 1 mm, this effect appears to have a limited impact on overall prediction accuracy.

The elevated Z-direction errors for central markers (\#3, 4, 5) reflect asymmetric indentation depth constraints in the experimental setup, where deeper indentation is possible at the centre than at the edges. This systematic bias may be reduced in future work through targeted data augmentation or by explicitly modelling position-dependent depth constraints during training.

\subsection{Comparison to Alternative Approaches}
The GNN achieves sub-millimetre prediction error with millisecond-scale inference, making it substantially faster than direct FEM while retaining sufficient accuracy (0.4 mm) for real-time visualisation and rapid design exploration.
Whereas the MuJoCo integration is used here for real-time visualisation, deformation prediction is provided by the learned surrogate trained on FEM-generated data, allowing hyperelastic behaviour to be approximated without running FEM online. Compared to standard data-driven methods that require thousands of real samples, our framework needs only minimal physical validation, making it especially suitable for fragile systems. More broadly, the same validation-then-surrogate strategy may help reduce calibration effort in other tactile simulation pipelines that rely on substantial real-world data.

\subsection{Limitations}
While our approach is effective, several areas present opportunities for future enhancement. The use of a planar 2D sensor simplifies the full 3D tactile deformation problem; extension to volumetric sensors will require 3D FEM meshes, volumetric graph representations, and additional computational considerations. For TPU material modeling, literature Ogden parameters proved generally sufficient, though sensor-specific testing could refine results for outliers. The current focus on quasi-static deformation could be expanded to capture dynamic effects like viscoelasticity or impact. Accuracy varies across the sensor, particularly for central markers and extended pins, suggesting value in position-aware modeling. Our MuJoCo integration currently supports visualization, and future development could add physics-based collision and force feedback for closed-loop control.

\section{Conclusions and Future Work}

We developed a data-efficient framework for planar 2D soft tactile sensor modelling, achieving sub-millimetre accuracy with only 82 physical measurements and synthetic training data. This approach enables robust sim-to-real transfer and real-time performance while dramatically reducing the need for costly experimentation, suggesting potential applicability to other fragile systems where large-scale testing is impractical. 

Future work will focus on extending this methodology to 3D sensor geometries, which introduce challenges such as volumetric mesh representation, more complex graphs, and increased computational cost, yet still require far fewer real measurements than traditional methods. We also aim to move beyond deformation prediction to force estimation and full physics integration, enabling grasp assessment and closed-loop tactile control. Expanding to multi-material and multi-contact scenarios, and leveraging active or transfer learning across sensor designs, could further reduce data requirements and broaden applicability. Together, these directions will advance scalable, accurate modelling for next-generation soft robotic sensors.
\noindent


\bibliographystyle{IEEEtran}
\bibliography{ref.bib}

@article{goury2018fast,
  title={Fast, generic, and reliable control and simulation of soft robots using model order reduction},
  author={Goury, Olivier and Duriez, Christian},
  journal={IEEE Transactions on Robotics},
  volume={34},
  number={6},
  pages={1565--1576},
  year={2018},
  publisher={IEEE}
}

@article{zhang2026simtac,
  title={SimTac: A Physics-Based Simulator for Vision-Based Tactile Sensing with Biomorphic Structures},
  author={Zhang, Xuyang and Jiang, Jiaqi and Chen, Zhuo and Zhao, Yongqiang and Yang, Tianqi and Gomes, Daniel Fernandes and Wang, Jianan and Luo, Shan},
  journal={Cyborg and Bionic Systems},
  volume={7},
  pages={0510},
  year={2026},
  publisher={AAAS}
}

@article{shi2024robocraft,
  title={Robocraft: Learning to see, simulate, and shape elasto-plastic objects in 3d with graph networks},
  author={Shi, Haochen and Xu, Huazhe and Huang, Zhiao and Li, Yunzhu and Wu, Jiajun},
  journal={The International Journal of Robotics Research},
  volume={43},
  number={4},
  pages={533--549},
  year={2024},
  publisher={SAGE Publications Sage UK: London, England}
}

@article{zhang2026embodied,
  title={Embodied Intelligence Technology for Next-Generation Soft Robots: A Survey},
  author={Zhang, Jingyu and Shen, Chenyang and Zhu, Zixin and Tong, Haochuan and Fang, Qin},
  journal={Engineered Science},
  volume={40},
  pages={2073},
  year={2026},
  publisher={Engineered Science Publisher}
}

@inproceedings{peng2018sim,
  title={Sim-to-real transfer of robotic control with dynamics randomization},
  author={Peng, Xue Bin and Andrychowicz, Marcin and Zaremba, Wojciech and Abbeel, Pieter},
  booktitle={2018 IEEE international conference on robotics and automation (ICRA)},
  pages={3803--3810},
  year={2018},
  organization={IEEE}
}

@article{thuruthel2018model,
  title={Model-based reinforcement learning for closed-loop dynamic control of soft robotic manipulators},
  author={Thuruthel, Thomas George and Falotico, Egidio and Renda, Federico and Laschi, Cecilia},
  journal={IEEE Transactions on Robotics},
  volume={35},
  number={1},
  pages={124--134},
  year={2018},
  publisher={IEEE}
}

@inproceedings{ajay2018augmenting,
  title={Augmenting physical simulators with stochastic neural networks: Case study of planar pushing and bouncing},
  author={Ajay, Anurag and Wu, Jiajun and Fazeli, Nima and Bauza, Maria and Kaelbling, Leslie P and Tenenbaum, Joshua B and Rodriguez, Alberto},
  booktitle={2018 IEEE/RSJ International Conference on Intelligent Robots and Systems (IROS)},
  pages={3066--3073},
  year={2018},
  organization={IEEE}
}

@inproceedings{sanchez2018graph,
  title={Graph networks as learnable physics engines for inference and control},
  author={Sanchez-Gonzalez, Alvaro and Heess, Nicolas and Springenberg, Jost Tobias and Merel, Josh and Riedmiller, Martin and Hadsell, Raia and Battaglia, Peter},
  booktitle={International conference on machine learning},
  pages={4470--4479},
  year={2018},
  organization={PMLR}
}

@inproceedings{matas2018sim,
  title={Sim-to-real reinforcement learning for deformable object manipulation},
  author={Matas, Jan and James, Stephen and Davison, Andrew J},
  booktitle={Conference on Robot Learning},
  pages={734--743},
  year={2018},
  organization={PMLR}
}

@article{battaglia2018relational,
  title={Relational inductive biases, deep learning, and graph networks},
  author={Battaglia, Peter W and Hamrick, Jessica B and Bapst, Victor and Sanchez-Gonzalez, Alvaro and Zambaldi, Vinicius and Malinowski, Mateusz and Tacchetti, Andrea and Raposo, David and Santoro, Adam and Faulkner, Ryan and others},
  journal={arXiv preprint arXiv:1806.01261},
  year={2018}
}

@inproceedings{sanchez2020learning,
  title={Learning to simulate complex physics with graph networks},
  author={Sanchez-Gonzalez, Alvaro and Godwin, Jonathan and Pfaff, Tobias and Ying, Rex and Leskovec, Jure and Battaglia, Peter},
  booktitle={International conference on machine learning},
  pages={8459--8468},
  year={2020},
  organization={PMLR}
}

@inproceedings{pfaff2020learning,
  title={Learning mesh-based simulation with graph networks},
  author={Pfaff, Tobias and Fortunato, Meire and Sanchez-Gonzalez, Alvaro and Battaglia, Peter},
  booktitle={International conference on learning representations},
  year={2020}
}

@inproceedings{duret2026real,
  title={Real-Time Simulation of Deformable Tactile Sensors and Objects in Robotic Grasping using Graph Neural Networks with Inductive Biases},
  author={Duret, Guillaume and Heller, Frederik and Mazurak, Danylo and Kshirsagar, Alap and Schneider, Tim and Zara, Florence and Peters, Jan and Chen, Liming},
  booktitle={9th IEEE-RAS International Conference on Soft Robotics (RoboSoft 2026)},
  year={2026}
}

@article{qin2024modeling,
  title={Modeling and simulation of dynamics in soft robotics: A review of numerical approaches},
  author={Qin, Longhui and Peng, Haijun and Huang, Xiaonan and Liu, Mingchao and Huang, Weicheng},
  journal={Current Robotics Reports},
  volume={5},
  number={1},
  pages={1--13},
  year={2024},
  publisher={Springer}
}

@article{yuan2017gelsight,
  title={Gelsight: High-resolution robot tactile sensors for estimating geometry and force},
  author={Yuan, Wenzhen and Dong, Siyuan and Adelson, Edward H},
  journal={Sensors},
  volume={17},
  number={12},
  pages={2762},
  year={2017},
  publisher={MDPI}
}

@article{lambeta2020digit,
  title={Digit: A novel design for a low-cost compact high-resolution tactile sensor with application to in-hand manipulation},
  author={Lambeta, Mike and Chou, Po-Wei and Tian, Stephen and Yang, Brian and Maloon, Benjamin and Most, Victoria Rose and Stroud, Dave and Santos, Raymond and Byagowi, Ahmad and Kammerer, Gregg and others},
  journal={IEEE Robotics and Automation Letters},
  volume={5},
  number={3},
  pages={3838--3845},
  year={2020},
  publisher={IEEE}
}

@inproceedings{kikuuwe2004tactile,
  title={The tactile contact lens},
  author={Kikuuwe, Ryo and Sano, Akihito and Mochiyama, Hiromi and Takesue, Naoyuki and Tsunekawa, K and Suzuki, S and Fujimoto, Hideo},
  booktitle={SENSORS, 2004 IEEE},
  pages={535--538},
  year={2004},
  organization={IEEE}
}

@article{martin2025anisotropic,
  title={Anisotropic and hyperelastic mechanical response of 3D printed TPU parts},
  author={Mart{\'\i}n-Sosa, Ezequiel and T{\'a}vara, Luis and Ojeda, Joaqu{\'\i}n and Estefani, Alejandro},
  journal={Progress in Additive Manufacturing},
  volume={10},
  number={8},
  pages={5697--5709},
  year={2025},
  publisher={Springer}
}

@article{smith2024mechanical,
  title={Mechanical characterization and constitutive modeling of 3D printable soft materials},
  author={Smith, Lawrence and MacCurdy, Robert},
  journal={3D Printing and Additive Manufacturing},
  volume={11},
  number={3},
  pages={1209--1212},
  year={2024},
  publisher={SAGE Publications Sage CA: Los Angeles, CA}
}

@inproceedings{todorov2012mujoco,
  title={Mujoco: A physics engine for model-based control},
  author={Todorov, Emanuel and Erez, Tom and Tassa, Yuval},
  booktitle={2012 IEEE/RSJ international conference on intelligent robots and systems},
  pages={5026--5033},
  year={2012},
  organization={IEEE}
}

@incollection{coumans2015bullet,
  title={Bullet physics simulation},
  author={Coumans, Erwin},
  booktitle={ACM SIGGRAPH 2015 Courses},
  pages={1},
  year={2015}
}

@article{makoviychuk2021isaac,
  title={Isaac gym: High performance gpu-based physics simulation for robot learning},
  author={Makoviychuk, Viktor and Wawrzyniak, Lukasz and Guo, Yunrong and Lu, Michelle and Storey, Kier and Macklin, Miles and Hoeller, David and Rudin, Nikita and Allshire, Arthur and Handa, Ankur and others},
  journal={arXiv preprint arXiv:2108.10470},
  year={2021}
}

@incollection{duriez2016soft,
  title={Soft robot modeling, simulation and control in real-time},
  author={Duriez, Christian and Bieze, Thor},
  booktitle={Soft Robotics: Trends, Applications and Challenges: Proceedings of the Soft Robotics Week, April 25-30, 2016, Livorno, Italy},
  pages={103--109},
  year={2016},
  publisher={Springer}
}

@article{ward2018tactip,
  title={The tactip family: Soft optical tactile sensors with 3d-printed biomimetic morphologies},
  author={Ward-Cherrier, Benjamin and Pestell, Nicholas and Cramphorn, Luke and Winstone, Benjamin and Giannaccini, Maria Elena and Rossiter, Jonathan and Lepora, Nathan F},
  journal={Soft robotics},
  volume={5},
  number={2},
  pages={216--227},
  year={2018},
  publisher={SAGE Publications Sage CA: Los Angeles, CA}
}

@inproceedings{tobin2017domain,
  title={Domain randomization for transferring deep neural networks from simulation to the real world},
  author={Tobin, Josh and Fong, Rachel and Ray, Alex and Schneider, Jonas and Zaremba, Wojciech and Abbeel, Pieter},
  booktitle={2017 IEEE/RSJ international conference on intelligent robots and systems (IROS)},
  pages={23--30},
  year={2017},
  organization={IEEE}
}

@article{wang2019dynamic,
  title={Dynamic graph cnn for learning on point clouds},
  author={Wang, Yue and Sun, Yongbin and Liu, Ziwei and Sarma, Sanjay E and Bronstein, Michael M and Solomon, Justin M},
  journal={ACM Transactions on Graphics (tog)},
  volume={38},
  number={5},
  pages={1--12},
  year={2019},
  publisher={Acm New York, NY, USA}
}

@article{lin2023comparative,
  title={Comparative analysis of various hyperelastic models and element types for finite element analysis},
  author={Lin, Po-Sen and Le Roux de Bretagne, Olivier and Grasso, Marzio and Brighton, James and StLeger-Harris, Chris and Carless, Owen},
  journal={Designs},
  volume={7},
  number={6},
  pages={135},
  year={2023},
  publisher={MDPI}
}

\end{document}